\documentclass{article} 
\usepackage[final]{neurips_2025} 

\PassOptionsToPackage{table}{xcolor}

\usepackage{booktabs}
\usepackage{longtable}
\usepackage{array}
\usepackage{colortbl}
\usepackage{multicol}
\usepackage{multirow}
\usepackage{threeparttable}

\usepackage{caption}
\usepackage{subcaption}
\usepackage{listings}
\usepackage{xcolor}

\lstdefinelanguage{json}{
    basicstyle=\ttfamily\small,
    numbers=left,
    numberstyle=\tiny\color{gray},
    stepnumber=1,
    numbersep=6pt,
    showstringspaces=false,
    breaklines=true,
    frame=single,
    backgroundcolor=\color{gray!5},
}

\usepackage[utf8]{inputenc}
\usepackage[T1]{fontenc}

\usepackage{hyperref}
\hypersetup{
  colorlinks=true,
  linkcolor=red,
  citecolor=cyan,
  filecolor=magenta,
  urlcolor=magenta,
}
\usepackage{url}

\usepackage{amsfonts}
\usepackage{amsmath}
\usepackage{amssymb}
\usepackage{nicefrac}

\usepackage{microtype}
\usepackage{xspace}
\usepackage{enumitem}
\usepackage{textcomp}
\usepackage{stfloats}
\usepackage{verbatim}
\usepackage{wrapfig}
\usepackage{graphicx}

\usepackage{cite}

\usepackage{tikz}
\usepackage{algorithm}
\usepackage{algpseudocode}
\usepackage{pgfplots}
\pgfplotsset{compat=1.18}

\title{Matrix-Game 3.0: Real-Time and Streaming Interactive World Model with Long-Horizon Memory}
\vspace{2em}
\author{%
  \textbf{Zile Wang}\thanks{Equal contribution.}\quad
  \textbf{Zexiang Liu}$^{*}$\quad
  \textbf{Jiaxing Li}$^{*}$\quad
  \textbf{Kaichen Huang}$^{*}$\quad
  \textbf{Baixin Xu}$^{*}$ \\
  \textbf{Fei Kang}\quad
  \textbf{Mengyin An}\quad
  \textbf{Peiyu Wang}\quad
  \textbf{Biao Jiang}\quad
  \textbf{Yichen Wei}\quad
  \textbf{Yidan Xietian}\\
  \textbf{Jiangbo Pei}\quad
  \textbf{Liang Hu}\quad
  \textbf{Boyi Jiang}\quad
  \textbf{Hua Xue}\quad
  \textbf{Zidong Wang}\quad
  \textbf{Haofeng Sun}\\
  \textbf{Wei Li}\quad
  \textbf{Wanli Ouyang}\quad
  \textbf{Xianglong He}\thanks{Project lead.}\quad 
  \textbf{Yang Liu}$^{\dag\ddag}$\quad
  \textbf{Yangguang Li}$^{\dag}$\thanks{Corresponding author: \texttt{matrix@email.skywork.ai}}\quad
  \textbf{Yahui Zhou} \\
  \\
  Skywork AI \\
  Project page: \href{https://matrix-game-v3.github.io/}{Matrix-Game-3.0-Homepage}
}

\begin{document}
\maketitle
\vspace{-6mm}
\begin{figure}[H]
    \centering
    \includegraphics[width=\linewidth]{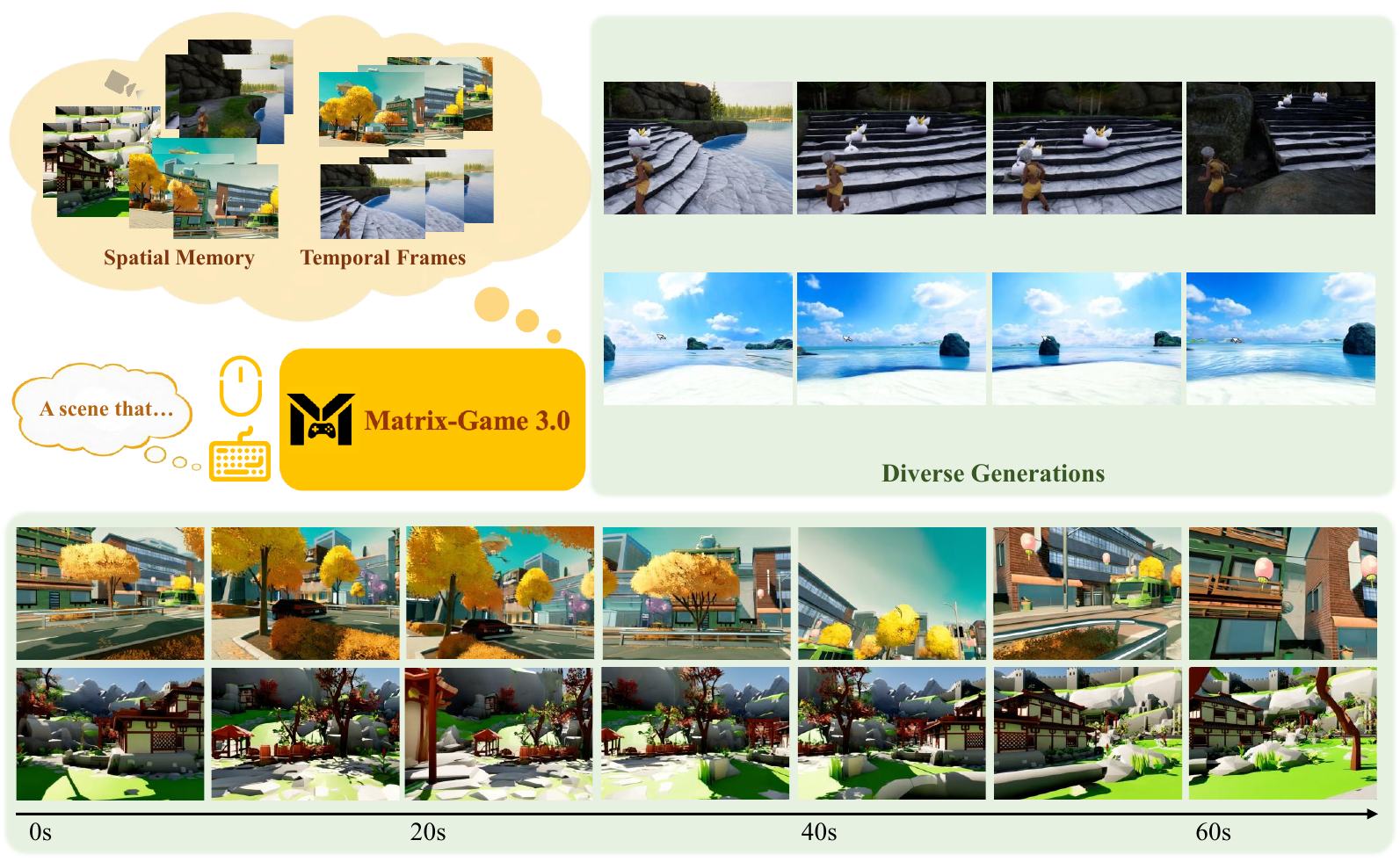}
    \caption{\textbf{Matrix-Game 3.0} introduces precise action control and long-horizon memory retrieval, enabling an interactive world model with long-term memory and real-time performance of up to 40 FPS.}
    \label{fig:teaser}
    \vspace{-5mm}
\end{figure}

\begin{abstract}
With the advancement of interactive video generation, diffusion models have increasingly demonstrated their potential as world models. However, existing approaches still struggle to simultaneously achieve memory-enabled long-term temporal consistency and high-resolution real-time generation, limiting their applicability in real-world scenarios. To address this, we present \textbf{Matrix-Game 3.0}, a memory-augmented interactive world model designed for 720p real-time long-form video generation.
Building upon Matrix-Game 2.0, we introduce systematic improvements across data, model, and inference. \textbf{First}, we develop an upgraded industrial-scale infinite data engine that integrates Unreal Engine--based synthetic data, large-scale automated collection from AAA games, and real-world video augmentation to produce high-quality Video--Pose--Action--Prompt quadruplet data at scale.
\textbf{Second}, we propose a training framework for long-horizon consistency: by modeling prediction residuals and re-injecting imperfect generated frames during training, the base model learns self-correction;
meanwhile, camera-aware memory retrieval and injection enable the base model to achieve long horizon spatiotemporal consistency.
\textbf{Third}, we design a multi-segment autoregressive distillation strategy based on Distribution Matching Distillation (DMD), combined with model quantization and VAE decoder pruning, to achieve efficient real-time inference.
Experimental results show that Matrix-Game 3.0 achieves up to 40 FPS real-time generation at 720p resolution with a 5B model, while maintaining stable memory consistency over minute-long sequences. Scaling up to a 2$\times$14B model further improves generation quality, dynamics, and generalization. Our approach provides a practical pathway toward industrial-scale deployable world models.
\end{abstract}

\section{Introduction}
Building world models to simulate complex environment dynamics and predict future observations under user actions, has attracted intense recent attention~\cite{genie3, dreamzero, vjepa2}. These models offer broad applicability across domains such as robotics planning and control~\cite{cosmospolicy, dreamzero}, entertainment~\cite{team2026advancing, gamecraft2, feng2024matrix, zhang2025matrixgame}, and interactive experiences in extended reality (XR)~\cite{worldlabs2025marble, yang2025matrix, team2025hunyuanworld}.
With the striking progress in diffusion-based short video generation~\cite{wan2025wan, hacohen2026ltx, kong2024hunyuanvideo} over the past few years, video models are increasingly recognized for their vast potential as world simulators. While demonstrating the ability to synthesize high-resolution, temporally coherent clips at scale, these models also encode an understanding of world knowledge and the capacity to predict future observations, thereby enabling rapid adaptation to a broad range of promising world model applications~\cite{gao2026dreamdojo, hu2025astranav}. For instance, interactive entertainment and gaming require persistent worlds that endure throughout exploration; open-ended interaction demands long-term memory of previously occurred events; and embodied intelligence and industrial workflows call for fine-grained, generalizable control aligned with real-world behavior~\cite{zhang2025world, hafner2025training, mei2026video}.

However, a common prerequisite across these scenarios is \textbf{real-time generation with long-horizon spatiotemporal consistency}: the ability to generate content continuously at real-time speed while preserving semantic and geometric coherence over extended horizons. Current powerful short-video diffusion models, when directly applied to such settings, lack this critical capability, that serves as a foundational requirement for world models. Without this foundation, downstream systems either degrade into incoherent short segments or incur prohibitive latency, making reliable deployment as world simulators in practical scenarios difficult.

A number of exploratory efforts of world models~\cite{bruce2024genie, parkerholder2024genie2, hong2025relic, mao2025yume, yu2026mosaicmem, zhang2026astrolabe, wang2026worldcompass, chen2025deepverse} have demonstrated the ability to generate longer sequences, yet providing accurate and stable future predictions remains under-explored. Several approaches have nonetheless shown the effectiveness of leveraging video generation models for interactive world simulation:
Genie~3~\cite{genie3} demonstrates that learning interaction-conditioned world dynamics from large-scale, long-temporal visual data is feasible and scalable, suggesting that practical interactive world modeling is increasingly within reach.
At the same time, the detailed training recipes, compute budgets, and end-to-end inference stacks behind such systems are not publicly released, limiting reproducibility and making it difficult for the research community to isolate which design choices drive long-horizon stability versus latency.
Accordingly, a growing body of public studies~\cite{oasis2024,hong2025relic,he2025matrix} targets controllable and interactive world models, yet simultaneously satisfying long-horizon memory consistency, high-resolution fidelity, and true real-time interaction remains rare in open works.
Prior works such as Matrix-Game~2.0~\cite{he2025matrix} and HY-Gamecraft-2~\cite{tang2025hunyuan} achieve real-time streaming interactive generation via causal autoregressive few-step diffusion, but lack memory mechanisms for stable minute-long consistency; Lingbot-World~\cite{team2026advancing} improves long-horizon geometric consistency by scaling context length of the diffusion model, yet simultaneously maintaining memory capability and robust real-time streaming deployment remains challenging.

\begin{figure}[t]
    \centering
    \includegraphics[width=1\linewidth]{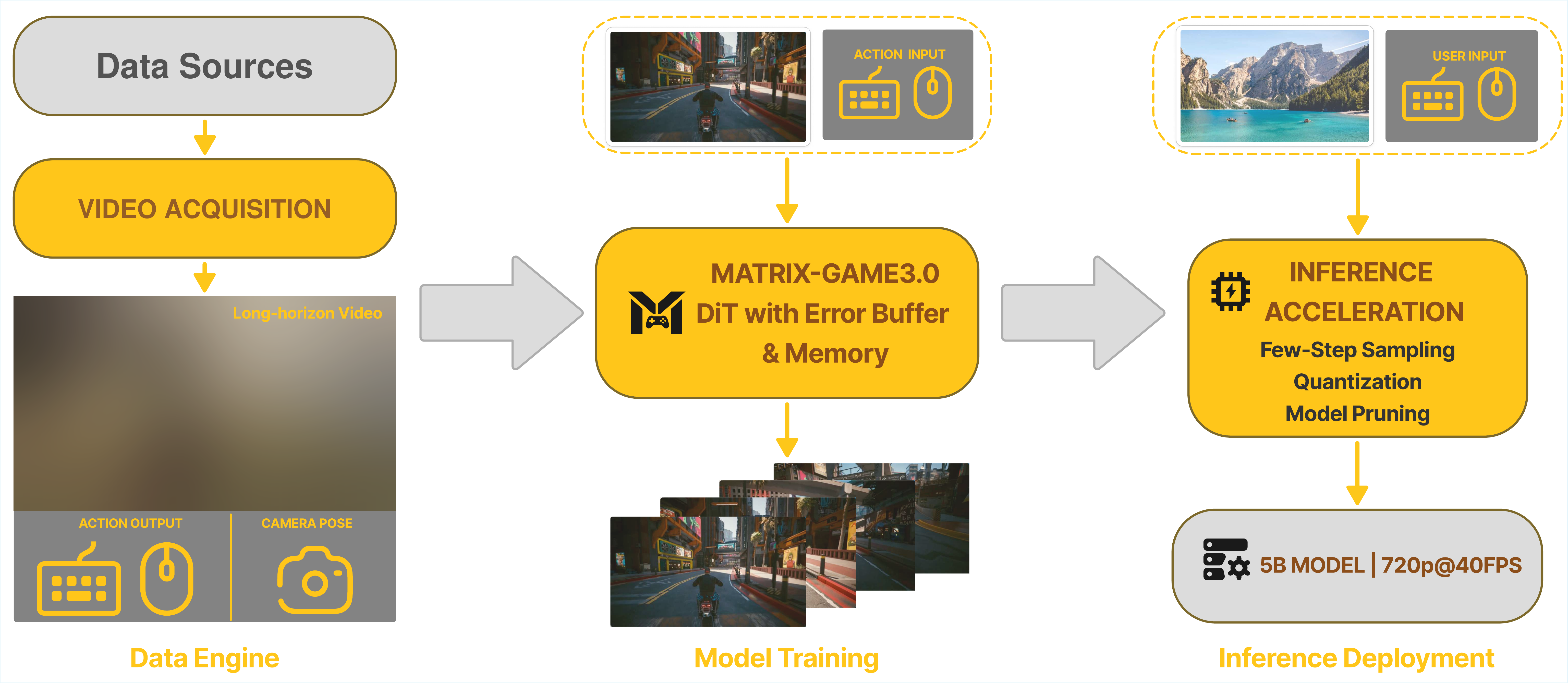}
    \caption{Overview of \textbf{Matrix-Game 3.0}. Our framework combines synthetic rendering, game-environment recording, and real-world video data with memory-augmented DiT training and accelerated real-time deployment. It prepares long-horizon training videos with action and camera-pose annotations, learns action-conditioned generation with memory-enhanced consistency, and supports real-time inference through few-step sampling, quantization, and pruning, achieving 720p@40FPS with a 5B model. The illustrative image at the lower left of the data-engine panel is blurred for publication-related redaction.}
    \label{fig:mg3_overview}
\end{figure}

To this end, we present \textbf{Matrix-Game~3.0}, as shown in Figure~\ref{fig:mg3_overview}, which reaches up to 40~FPS at 720p with a 5B model while maintaining stable memory consistency over minute-long sequences; scaling to 28B further improves generation quality, dynamics, and generalization.

Specifically, turning such capabilities into practice calls for coordinated advances along three tightly coupled factors, which motivate a co-designed solution across \emph{data}, \emph{modeling}, and \emph{deployment}.

\paragraph{Data.} Genie-3 demonstrates that interactive controllability and memory capability can be learned from precisely annotated large-scale video datasets. However, obtaining synchronized action and camera-pose supervision together with broad scene coverage remains challenging for a single data source.
In that case, \textbf{we develop a data preparation system that integrates three complementary sources}: (i) an Unreal Engine~5 synthetic pipeline with tick-synchronized video from navigation-mesh-based exploration, stochastic camera control, and a combinatorial character assembly system yielding over $10^8$ variants; (ii) a four-layer decoupled architecture for automated recording of gameplay video and associated state information; and (iii) diverse real-world corpora (DL3DV-10K~\cite{ling2024dl3dv}, RealEstate10K~\cite{zhou2018stereo}, OmniWorld~\cite{zhou2025omniworld}, and SpatialVid~\cite{wang2025spatialvid}) spanning indoor, urban, aerial, and vehicular scenes.
Together, these sources provide annotated video data for training long interactive rollouts.

\paragraph{Modeling.}
Bridging strong bidirectional video priors with a streaming inference paradigm introduces inherent trade-offs among long-horizon memory, coherence, controllability, and error accumulation. Prior  works explore complementary strategies: 
existing distillation-based approaches (e.g., HY-World, RELIC) typically transfer memory and controllability from bidirectional base models to causal few-step models. However, they largely overlook the limited robustness of the base models to error accumulation, as well as the distribution gap between the student and the base model;
scaling-based approaches (e.g., Lingbot-World) extend context length for improved long-term consistency, yet remaining such capacity with real-time inference deployment is nontrivial. 
To address these challenges, \textbf{we adopt a bidirectional backbone with a camera-aware memory retrieval mechanism}: we aim to preserve the inherent strengths of bidirectional model priors while retrieving injecting memory for long-horizon spatiotemporal consistency.  We additionally introduce error-aware training~\cite{li2025stable} to learn self-correction for the base model, better aligning with multi-segment generation process.

\paragraph{Deployment.}
Industrial system suggests that real-time high-resolution interaction is achievable~\cite{genie3}, but training recipes and full inference pipelines are often undisclosed. To this end, \textbf{we introduce a multi-segment distillation method for bidirectional models}, inspired by Distribution Matching Distillation (DMD)~\cite{yin2024one} and Self-Forcing~\cite{huang2025self} paradigms, reducing error accumulation yet achieving streaming inference. \textbf{We further deploy a series of acceleration techniques to achieve 40FPS generaiton at 720p resolution for 5B parameter model.} (e.g. DiT quantization, VAE pruning, retrieval via GPU, etc.)

\section{Related Works}
\subsection{Video Generation Models}
Recent video generation models have largely converged toward Diffusion Transformer (DiT)-based architectures~\cite{peebles2023scalable}, which directly model spatiotemporal tokens and enable scalable high-resolution and high-quality video synthesis. Closed-source models such as Sora~\cite{openai2024worldsim}, Kling~\cite{team2025kling}, and Hailuo have achieved significant progress in complex motion modeling and high-quality video generation through large-scale data and model scaling. However, these approaches are primarily designed for offline generation and lack explicit modeling of actions and interaction. Moreover, their long-horizon consistency typically relies on implicit modeling mechanisms, making it difficult to maintain stable performance in extended sequence generation.
In parallel, open-source models (e.g., Wan~\cite{wan2025wan}, Magi-1~\cite{teng2025magi}, and LTX-2.3~\cite{hacohen2026ltx}) aim to advance video generation research by improving openness and scalability. In particular, Wan~\cite{wan2025wan} builds a DiT-based video generation system enhanced by large-scale data and training strategies; Magi-1~\cite{teng2025magi} adopts a chunk-wise autoregressive diffusion paradigm, decomposing videos into sequential segments to enable scalable long-horizon modeling; and LTX-2.3~\cite{hacohen2026ltx} further extends this line to audio-visual generation, supporting synchronized video and audio generation within a unified framework, thereby enabling joint modeling of visual and acoustic signals. Despite these advances, these models remain largely focused on offline generation, and their long-horizon consistency relies on implicit mechanisms, limiting their applicability in interactive or long sequence scenarios.

\subsection{Long-Horizon Video Generation}
Long-horizon video generation is fundamentally constrained by error accumulation and training–inference mismatch in autoregressive generation.
From an architectural perspective, causal video diffusion models introduce causal attention and KV cache reuse to enable efficient autoregressive generation, but they do not fundamentally address error accumulation, where small prediction errors compound over time and lead to temporal drift. From a training perspective, Diffusion Forcing~\cite{chen2024diffusion} unifies next-step prediction and full-sequence diffusion by applying independent noise levels to tokens, enabling extrapolation beyond the training horizon and improving stability. However, it still relies on ground-truth or noisy ground-truth inputs during training, resulting in a mismatch between training and inference distributions.
To mitigate this, Self-Forcing~\cite{huang2025self} introduces autoregressive rollout during training, conditioning each frame on previously generated outputs to better simulate inference and reduce exposure bias. Building upon this, Self-Forcing++~\cite{cui2025self} extends the approach to minute-scale generation through mechanisms such as rolling KV cache and distribution matching, improving scalability. Furthermore, Causal Forcing~\cite{zhu2026causal} identifies the architectural mismatch between bidirectional teacher models and autoregressive students, and proposes causal teacher initialization to improve training–inference consistency and generation quality. In addition, SVI-style~\cite{li2025stable} approaches explicitly model prediction errors and incorporate feedback correction to enable self-correction under accumulated noise.
Overall, these methods improve long-horizon generation by mitigating exposure bias and training–inference mismatch, enhancing autoregressive stability, and reducing error accumulation. However, they still lack explicit memory mechanisms and struggle to maintain spatial consistency across viewpoints, especially under high-resolution real-time settings.

\subsection{Interactive World Models}
Interactive world models aim to extend video generation into modeling state–action–environment transitions, where future observations are conditioned not only on past visual context but also on external actions. Compared to passive video synthesis, this setting requires jointly achieving controllability, long-horizon consistency, and real-time inference, making it significantly more challenging.
A representative line of work is the Genie~\cite{bruce2024genie}, which demonstrates that action-conditioned world models can be learned from large-scale unlabeled videos without explicit action annotations. Genie-2~\cite{parkerholder2024genie2} further extends this paradigm to generate interactive 3D environments that can be explored through action control, improving both interaction capability and environment modeling. However, these approaches remain limited by short-term memory and exhibit instability in long-horizon generation. The latest Genie-3~\cite{genie3} enables real-time interactive world simulation, generating navigable environments at approximately 24 FPS and 720p resolution while maintaining coherence for minute level sequence. It also exhibits implicit memory and persistent spatial consistency, but is still not open source and the details are unclear. 
In open-source research, models such as OASIS~\cite{oasis2024}, Matrix-Game 2.0~\cite{he2025matrix}, and WorldPlay~\cite{sun2025worldplay} progressively push interactive world models toward unified modeling and real-time interaction. OASIS~\cite{oasis2024} integrates video generation, action conditioning, and multimodal control into a unified framework; Matrix-Game 2.0~\cite{he2025matrix} adopts causal autoregressive diffusion with few-step distillation to enable streaming real-time interactive video generation with fine-grained action control; and WorldPlay~\cite{sun2025worldplay} further introduces memory-augmented mechanisms on top of real-time generation to improve long-horizon geometric consistency and stable interaction. Despite these advances, a trade-off between long-horizon consistency and high-resolution real-time generation still remains.
To further improve long-horizon consistency, subsequent works introduce memory-augmented mechanisms. For example, RELIC~\cite{hong2025relic} leverages a camera-aware KV cache to store historical latent features and retrieve relevant past observations during generation, improving cross-temporal and cross-view consistency. Prior to this, other related works~\cite{yu2025context,xiao2025worldmem,li2025vmem} have also explored this idea with learned memory retrieval and memory banks to better model long-term dependencies. However, these approaches typically introduce additional computational overhead, making them difficult to scale to high-resolution real-time settings.
Overall, existing methods advance along three main directions: action-conditioned modeling, memory-augmented consistency, and scalable training. Nevertheless, none of them simultaneously achieve long-horizon memory consistency, high-resolution generation, and real-time interactive capability within a unified framework.
\section{Method}
We propose the Matrix-Game 3.0 technical framework to address the memory challenge in long-horizon generation for interactive world models, as well as real-time generation at higher resolutions with larger models (e.g., a 5B model at 720p). The system consists of four key components:
Error-aware interactive base model, which ensures accurate action controllability and anti-drift consistency during long-term generation (see Sec.~\ref{sec:base_model} for details);
Camera-aware long-horizon memory mechanism, which equips the base model with strong long-horizon memory capabilities (see Sec.~\ref{sec:memory} for details);
Training–inference aligned few-step distillation pipeline, designed to enable distilled model stable few-step long-horizon generation with memory (see Sec.~\ref{sec:distillation} for details);
Real-time inference acceleration module, which ensures that the distilled model achieves real-time inference speed (see Sec.~\ref{sec:real-time} for details);
The coordinated integration of these four components enables Matrix-Game 3.0 to achieve strong long-horizon memory consistency and high-resolution real-time generation with a 5B model. Furthermore, we scale the model up to 28B parameters, demonstrating improved dynamic behavior and strong generalization capabilities at larger model scales (see Sec.~\ref{sec:large_model} for details).

\subsection{Error-Aware Interactive Base Model}
\label{sec:base_model}
\begin{figure}
    \centering
    \vspace{-3pt}
    \includegraphics[width=1\linewidth]{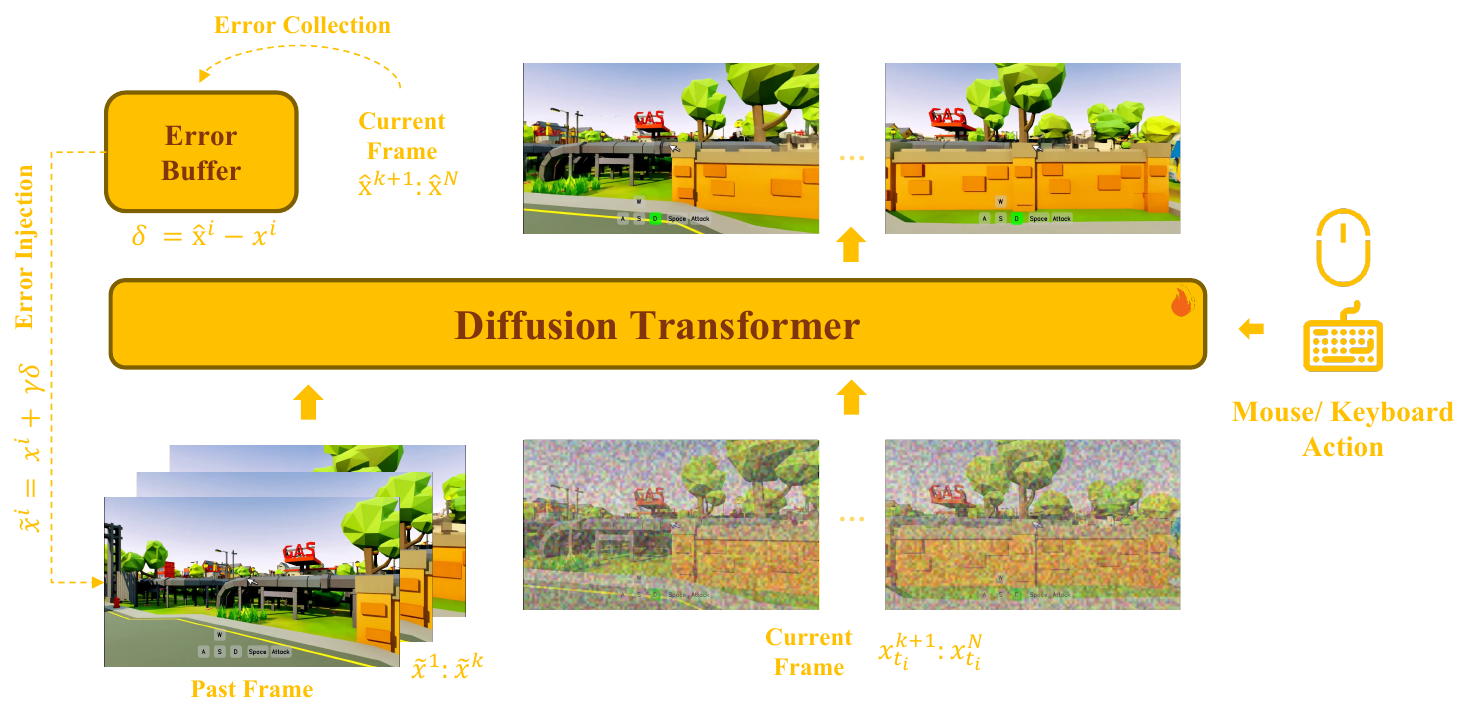}
    \caption{\textbf{Illustration of our interactive base model.} We jointly perform error-aware modeling over the past and current latent frames, while explicitly injecting action conditions into the model. This design enables autoregressive, long-horizon interactive generation and maintains consistency with the subsequent distillation stage.}
    \label{fig:base_model}
     \vspace{-3pt}
\end{figure}

Our pipeline is developed from an action-guided self-correcting base model. To maintain coherence and consistency with the subsequent distillation stage, we emphasize two key principles.

\textit{\textbf{(1) Architectural consistency.} Existing world-model\cite{he2025matrix} and video-generation\cite{huang2025self,zhu2026causal,yin2025causvid} methods commonly employ heterogeneous teacher–student architectures. However, theoretical studies\cite{zhu2026causal} suggest that such heterogeneity can lead to mismatched mappings and consequently unstable training. To address this issue, we use a unified bidirectional architecture for both the multi-step base model and the few-step distilled model. This unified design avoids the instability introduced by architectural heterogeneity, while also reducing the cost of ODE-based distillation.}

\textit{\textbf{(2) Robustness to imperfect contexts.} The behaviour of base model and distilled model should keep consistent when facing imperfect contexts (e.g. self-generated history latents during streaming inference), thus providing accurate distillation target and reducing exposure errors. Therefore, the base model should also be trained with imperfect historical contexts, rather than only clean ground-truth ones.}

Figure~\ref{fig:base_model} illustrates the design of our base model. Let $x^{1:N}$ denote a sequence of video latents. We partition the sequence into two groups: the first $k$ latent frames, referred to as \emph{past latent frames}, serve as the history condition, while the remaining $N-k$ latent frames, referred to as \emph{current latent frames}, correspond to the frames to be predicted by the model. Gaussian noise is randomly added to the current latent group, after which the two groups are concatenated and fed into a bidirectional diffusion Transformer. The flow-matching objective is imposed only on the current latent frames.

To enable precise action control, we follow the design of Matrix-Game 2.0~\cite{he2025matrix} and Game-Factory~\cite{yu2025gamefactory}, explicitly incorporating user actions into the model. Specifically, discrete keyboard actions are introduced through a dedicated Cross-Attention module, enabling accurate control over interactive behaviors. In contrast, continuous mouse-control signals are injected through Self-Attention, directly influencing the generation of the current visual state. This design enables stable and controllable interactive responses without sacrificing generation quality.

To enhance the long-horizon generation capability of the base model and equipping subsequent distillation stage, we follow SVI~\cite{li2025stable} to introduce error collection and error injection, simulating conditional contexts corrupted by exposure errors. Specifically, we maintain an error buffer $\mathcal{E}$ during training.
In the error collection stage, we first convert the model output into the corresponding clean estimate $\hat{x}^{i}$, i.e., the $x_0$ prediction implied by the predicted flow. The residual is then defined as
\begin{equation}
\delta = \hat{x}^{i} - x^{i}.
\end{equation}
All residuals are accumulated in the error buffer $\mathcal{E}$.
In the error injection stage, we sample $\delta \sim \mathrm{Uniform}(\mathcal{E})$ and perturb the history latent as
\begin{equation}
\tilde{x}^{i} = x^{i} + \gamma \delta,
\end{equation}
where $\gamma$ is a scalar controlling the perturbation magnitude.
The training objective is defined as
\begin{equation}
\mathcal{L}
=
\mathbb{E}_{x,t,\epsilon,\delta}
\left[
\left\|
\left(\epsilon-x^{k+1:N}\right)
-
v_\theta\!\left(x^{k+1:N}_{t},\, t \mid \tilde{x}^{1:k},c\right)
\right\|_2^2
\right].
\end{equation}
Here, $c$ denotes the action condition. This self-correcting formulation enables an interactive base model that supports autoregressive long-horizon generation and remains compatible with the downstream distillation process.

\subsection{Long-Horizon Memory}
\label{sec:memory}
\begin{figure}
    \centering
    \includegraphics[width=1.0\linewidth]{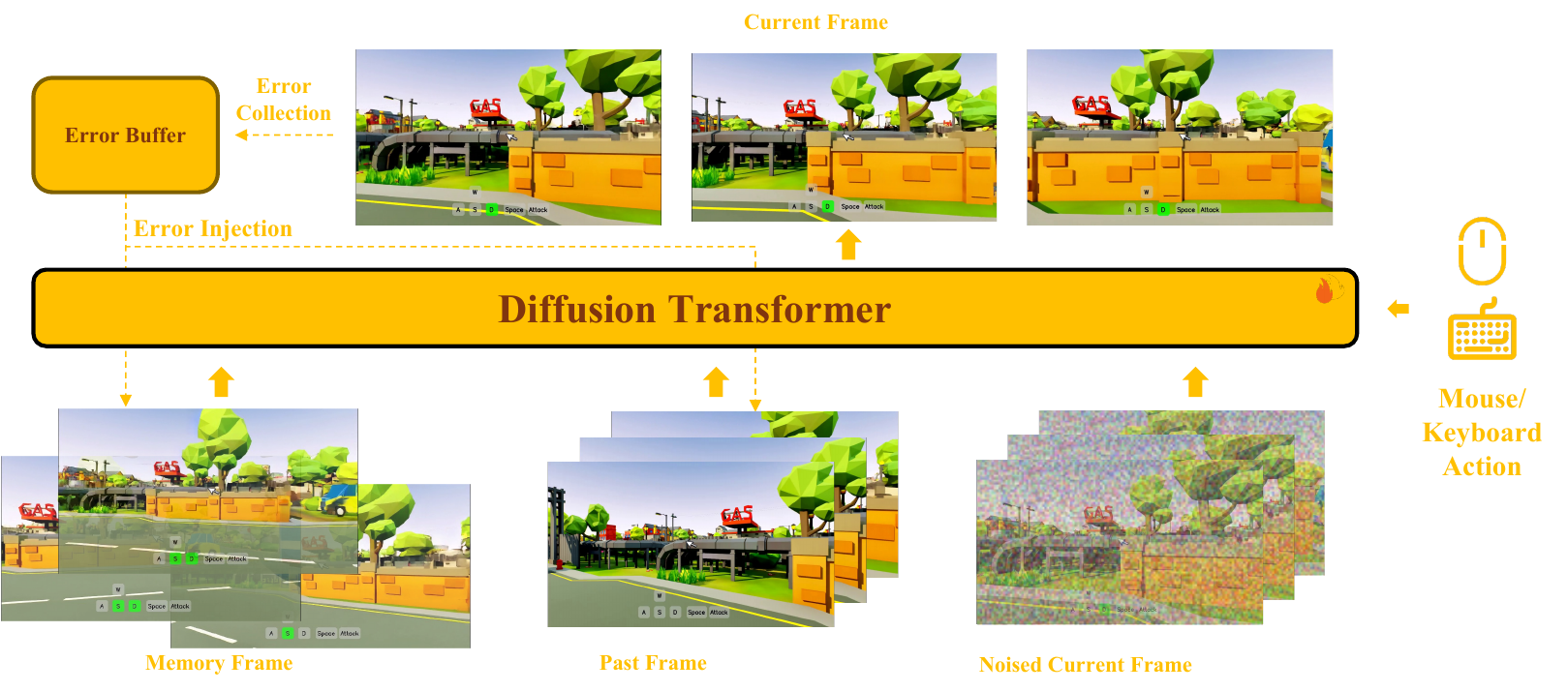}
    \caption{\textbf{Illustration of our memory-augmented base model.} Built upon the bidirectional base model, we incorporate retrieved memory frames as additional conditions and introduce small memory perturbations to enhance robustness. This design enables the base model to jointly model long-term memory, short-term history, and the current prediction target under the same attention mode as the base model.}
    \label{fig:memory_base_model}
\end{figure}

To endow a streaming interactive generator with memory, the current prediction should depend not only on temporally recent history latents, but also on earlier memory observations to keep the spatial consistency. We first examined two representative design directions.

\textit{
\textbf{(1) Implicit sparse long-context modeling.} Existing approaches such as MoC~\cite{cai2026moc} model long-horizon consistency by retrieving top-k similar chunks over a sparsely routed long context, which is conceptually appealing. However, during high-noise stages, unreliable similarity estimation leads to unstable memory selection, making convergence difficult for modeling long-term consistency under few-step settings. In addition, maintaining long-context sequences during training incurs substantial computational overhead.}

\textit{
\textbf{(2) Camera-aware explicit modeling.} A more straightforward approach is to retrieve memory frames based on camera awareness and inject them via cross-attention mechanism. Compared to MoC-style routing, this improves retrieval stability. However, the additional memory branch with misaligned features, together with layer-wise repeated feature injection, results in slow convergence. Even with geometry-aware cues inspired by prior memory-based world models~\cite{xiao2025worldmem}, the performance gains remain limited.}

\begin{wrapfigure}{r}{0.45\linewidth}
  \centering
  \vspace{-2em}
  \includegraphics[width=\linewidth]{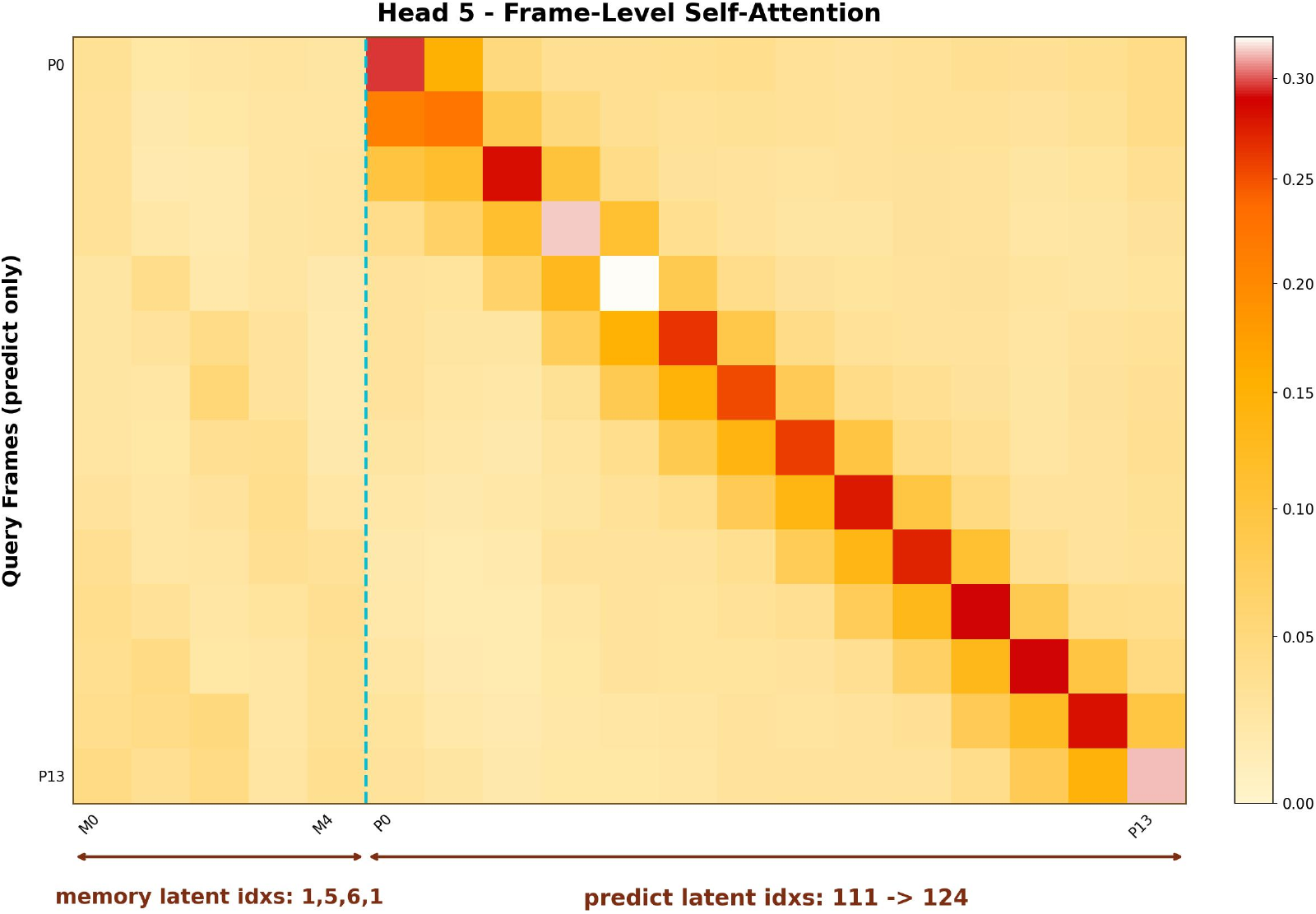}
  \vspace{-1em}
  \caption{Frame-level self-attention visualization for the memory-enhanced DiT.}
  \label{fig:memory_attn_vis}
  \vspace{-4mm}
\end{wrapfigure}

Based on these observations, we adopt a unified DiT framework that jointly models long-term memory, temporally consistent history, and the current prediction target.

Our first key design is a joint self-attention mechanism. Instead of treating memory as an external branch, we place retrieved memory latents, temporally aligned recent latents (where we also refer to \textit{past frames}), and current prediction noisy latents into the same attention space. In that case, the model can exchange information across both spatial and temporal levels within a single denoising hierarchy directly. As illustrated in Figure~\ref{fig:memory_base_model}, the retrieved memory, past frames, and noised current frames are jointly processed by the same Diffusion Transformer~\cite{peebles2023scalable} together with mouse/keyboard action conditions, while the model predicts only the current frames. The resulting prediction residuals are then collected into an error buffer and reinjected into the conditioning pathway during training, so that both history and memory conditions better match the imperfect contexts encountered during autoregressive inference. In this formulation, local history mainly supports short-term continuity, while retrieved memory provides longer-range anchors for scene layout, object state, and visual identity. This unified interaction is more compatible with streaming generation than maintaining a separate memory pathway, since memory and prediction features evolve together inside the same DiT backbone.

Our second key design is camera-aware memory selection together with relative Pl\"ucker encoding. Not all historical observations are useful for the current prediction, especially in long rollouts with frequent viewpoint changes. We therefore retrieve memory according to camera pose and field-of-view overlap, so that only view-relevant historical content is introduced into generation~\cite{yu2025contextmemory}. For inference, we optionally retain the first latent in the sequence as a persistent sink latent. This latent provides a stable global anchor for coarse scene style and appearance statistics across the full rollout, complementing the more view-specific retrieved memory frames. On top of retrieval, we explicitly encode the relative camera geometry between the current target and the selected memory through Pl\"ucker-style cues. This helps the model reason about how the same scene should be aligned across different viewpoints, and reduces the tendency to use historical information in a view-inconsistent manner.

To reduce the train-inference mismatch in the memory pathway, we introduce error collection and error injection for the full conditioning context, including both retrieved memory and recent history. During training, these conditioning latents are constructed from ground-truth video clips, whereas at inference time they are formed from previously generated frames and therefore inevitably contain accumulated errors. To bridge this gap, we maintain a shared latent error buffer \(\mathcal{E}\) for memory latents, recent past frame latents, and current prediction latents. In the error collection stage, we first convert the model output into the corresponding clean estimate \(\hat{z}^{i}\), namely the \(x_0\) prediction implied by the predicted flow, and define the residual as
\begin{equation}
\delta = \hat{z}^{i} - z^{i},
\end{equation}
where \(z^{i}\) denotes a latent sampled from memory, past frames, or current prediction. All residuals are accumulated in the shared latent error buffer \(\mathcal{E}\). In the error injection stage, we sample \(\delta \sim \mathrm{Uniform}(\mathcal{E})\) and perturb both the retrieved memory latents and the recent history latents as
\begin{equation}
\tilde{x}^{1:k} = x^{1:k} + \gamma_h \delta,\qquad
\tilde{m}^{1:r} = m^{1:r} + \gamma_m \delta,
\end{equation}
where \(x^{1:k}\) denotes the past latent frames, \(m^{1:r}\) denotes the retrieved memory latents, and \(\gamma_h\) and \(\gamma_m\) control the perturbation magnitudes for history and memory, respectively. The corresponding training objective is defined as
\begin{equation}
\mathcal{L}_{\mathrm{mem}}
=
\mathbb{E}_{x,m,t,\epsilon,\delta}
\left[
\left\|
\left(\epsilon-x^{k+1:N}\right)
-
v_\theta\!\left(x_t^{k+1:N},\, t \mid \tilde{x}^{1:k},\, \tilde{m}^{1:r},\, c,\, g\right)
\right\|_2^2
\right],
\end{equation}
where \(x^{k+1:N}\) denotes the current latent frames to be predicted, \(c\) denotes the action condition, and \(g\) denotes the geometric condition. This self-corrective formulation enables the model to learn how to extract useful information from imperfect memory and imperfect short-term history, thereby improving robustness under autoregressive rollout.

In addition, we strengthen temporal awareness in the rotary positional encoding. Besides relative offsets within the current segment, we inject the original frame index of each memory, history, and prediction latent into the temporal rotary construction, so that the model has access to their actual temporal locations in the full sequence. This helps the model distinguish recent history from truly distant memory and improves temporal disambiguation in long rollouts. At the same time, because RoPE is periodic, distant memory and current prediction frames may occasionally exhibit accidental positional alignment. To reduce this effect, we introduce a head-wise perturbed RoPE base during training. Concretely, for the \(h\)-th attention head, the effective rotary base is written as
\label{equation:LOL}
\begin{equation}
\hat{\theta}_h = \theta_{\mathrm{base}} \bigl(1 + \sigma_\theta \epsilon_h\bigr),
\end{equation}
where \(\epsilon_h\) is a head-dependent perturbation coefficient and \(\sigma_\theta\) controls the perturbation magnitude. In this way, different attention heads operate with different effective rotary bases, which helps break periodic synchronization across heads, mitigates positional aliasing, and discourages overly literal copying from temporally distant memory. As shown in Figure~\ref{fig:memory_attn_vis}, even when the selected memory frames are temporally far from the current prediction frames and therefore have substantially different rotary phases, their influence is not suppressed. The visualization averages self-attention over all DiT blocks, all denoising steps, and all within-frame tokens, and then aggregates the result to frame-level attention weights. Under this aggregation, the retrieved memory still receives clearly non-negligible attention, and in several cases its attention score remains comparable to nearby off-diagonal interactions among prediction frames. This suggests that the proposed temporal encoding design preserves useful long-range memory access while reducing periodic ambiguity.

Overall, the final memory design combines structured retrieval, unified self-attention, geometry-aware conditioning, and self-corrective memory training within a single DiT framework. 

\subsection{Training-Inference Aligned Few-step Distillation}
\label{sec:distillation}
\begin{figure}
    \centering
    \includegraphics[width=1\linewidth]{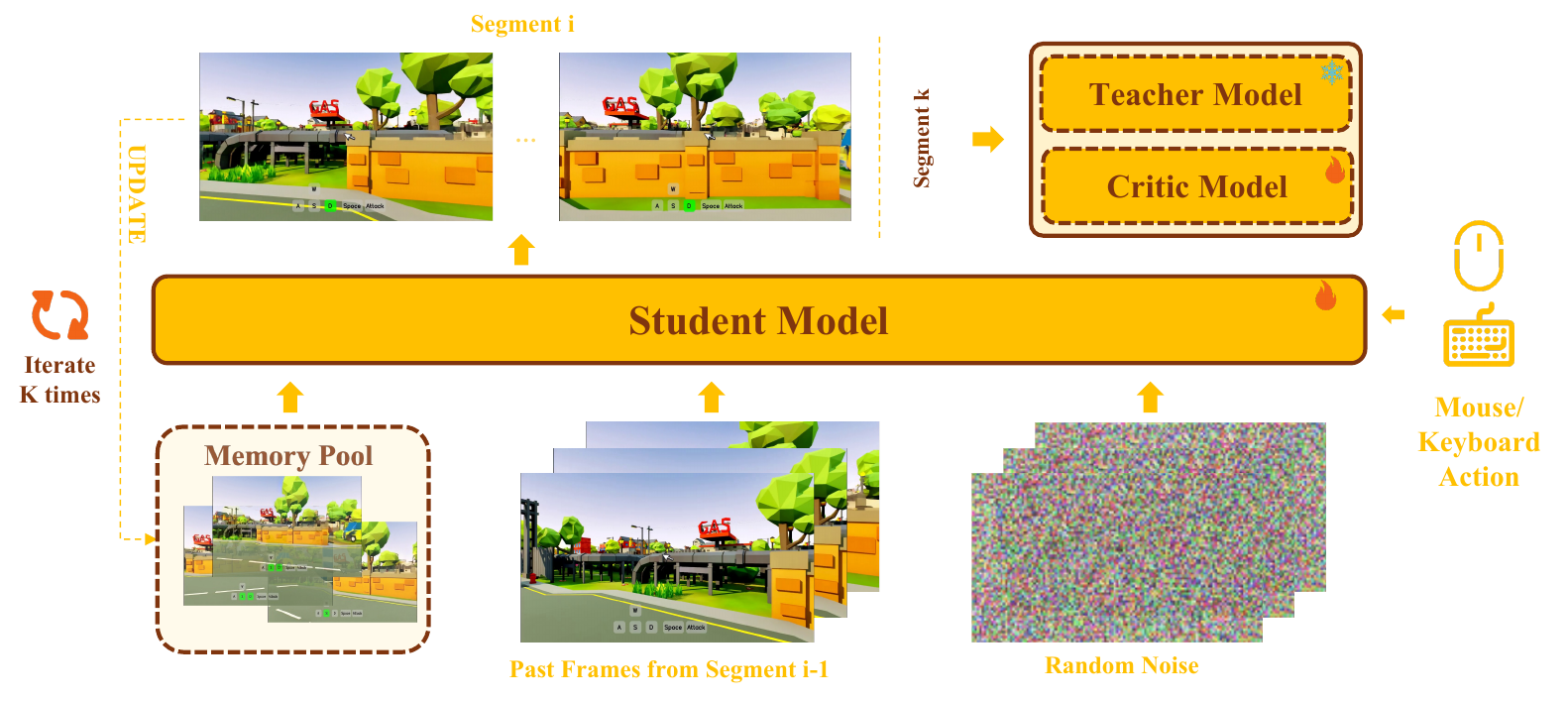}
    \caption{\textbf{Illustration of our few-step distillation stage.} The bidirectional student performs multi-segment rollouts to mimic actual few-step inference, with the final segment used for distribution matching, thereby ensuring training-inference consistency.}
    \label{fig:distill}
    \vspace{-1em}
\end{figure}

Existing distillation methods\cite{huang2025self,cui2025self,zhu2026causal,yin2025causvid} typically adopt causal students that perform chunk-wise inference, which naturally supports self-generated rollouts up to the length of a teacher window. In contrast, under bidirectional autoregressive modeling, the student inference span is no longer chunk-wise, but instead matches the teacher and covers the entire window. In this setting, single-window inference alone cannot provide the biased-but-clean history frames required for autoregressive generation. Naively using ground-truth frames as guidance creates a mismatch between training and actual inference, which in turn introduces exposure bias.

To ensure consistency between training and inference, we introduce a \emph{multi-segment self-generated inference scheme} for the bidirectional student, as illustrated in Figure~\ref{fig:distill}. Our method builds upon the idea of \emph{Distribution Matching Distillation (DMD)} for generation. Specifically, the bidirectional student is trained to mimic the actual few-step inference process by rolling out over multiple segments. Each segment starts from random noise. The past frames of the current segment are taken from the tail of the previous segment, while the memory signal is retrieved from an online-updated memory pool according to the current camera viewpoint. For the first segment, no memory is available, and the model operates in an image-to-video (I2V) mode. During training, we stop at a randomly selected segment, and feed the resulting segment to the teacher and critic. 

Next, the DMD objective minimizes the reverse KL divergence at sampled timesteps $t$ between the targeted data distribution $p_{\text{data}}(x^{\text{current}}_t)$ and the student distribution $p_{\text{gen}}(x^{\text{current}}_t)$. The gradient of the reverse KL can be approximated by the difference between two score functions:
\begin{align}
\nabla_{\theta} \mathcal{L}_{\text{DMD}}
&\triangleq
\mathbb{E}_{t}\!\left[\nabla_{\theta} D_{\mathrm{KL}}\!\left(p_{\theta,t}\,\|\,p_{\text{data},t}\right)\right] \notag \\
&\approx
-
\mathbb{E}_{t}\!\left[
\int
\left(
s_{\text{data}}(x^{\text{current}}_{t}, t, x^{\text{past}}, c, M)
-
s_{\text{gen},\xi}(x^{\text{current}}_{t}, t, x^{\text{past}}, c, M)
\right)
\nabla_{\theta} x_t \, d\epsilon
\right].
\label{eq:dmd}
\end{align}

Here, $c$ and $M$ denote the action condition and memory, respectively. $x^{\text{current}}$ corresponds to the current segment, while $x^{\text{past}}$ is taken from the end of the previous segment. With this multi-segment student inference scheme, we achieve few-step distillation with aligned training and inference behaviors, laying the foundation for real-time inference in the full pipeline.

\subsection{Real-Time Inference}
\label{sec:real-time}
Building upon the distilled model, we further adopt several strategies to accelerate inference\footnote{We also appreciate the support from Reactor (\url{https://www.reactor.inc}) in providing the insightful infrastructure for acceleration.}, including INT8 quantization for the DiT model, VAE pruning, and GPU-based memory retrieval. In our asynchronous deployment, 8 GPUs are used for DiT inference and 1 GPU is dedicated to VAE decoding, enabling the overall pipeline to achieve a inference speed of up to 40 FPS.

\textbf{DiT quantization.} To accelerate DiT inference, we apply INT8 quantization to the attention projection layers in DiT, while keeping the remaining components, including the FFN, VAE, and text encoder, in their original precision. This design reduces computation and memory overhead in the most critical part of the model, while maintaining overall generation quality. The underlying quantization operators are adapted from LightX2V~\cite{lightx2v}, providing an efficient implementation that accelerates inference without requiring full-model quantization.

\textbf{VAE pruning.} In the high-resolution streaming generation, VAE decoding becomes a major bottleneck. To address this, we train a lightweight VAE, denoted as \textbf{MG-LightVAE}. Following the homogeneous decoder design in LightX2V~\cite{lightx2v}, we decrease the hidden dimensions of the decoder while keeping the overall architecture unchanged. Our training pipeline is adapted from TurboVAED~\cite{zou2026turbo}, and the model is trained on 700K video clips of 17 frames, collected from a mixture of game environments and real-world scenes. 
We train two versions of MG-LightVAE, with 50\% and 75\% pruning ratios, achieving decoding speedups of $\times 2.6$ and $\times5.2$, respectively. 
In addition, we apply \texttt{torch.compile} to the VAE decoder after the first iteration to reduce decoding latency in subsequent steps. 

\textbf{Retrieval via GPU.} As described in this section, retrieving relevant memory information is a necessary step during generation, and we implement both CPU and GPU versions of camera-aware memory retrieval for this purpose. In practice, the GPU version significantly reduces retrieval time. 

At iteration $k$, each query view selects the past frame with the highest geometric overlap:
\begin{equation}
j_i^\star = \arg\max_{j \in \mathcal{C}_k} s(i,j).
\end{equation}

The CPU method uses the exact frustum-overlap score:
\begin{equation}
s_{\text{exact}}(i,j) = 
\frac{\operatorname{Vol}\big(\mathcal{F}(E_i) \cap \mathcal{F}(E_j)\big)}
{\operatorname{Vol}\big(\mathcal{F}(E_i)\big)},
\end{equation}
which is accurate but expensive.

In contrast, the GPU method uses a sampling-based approximation:
\begin{equation}
s_{\text{approx}}(i,j) = 
\frac{1}{N} \sum_{n=1}^{N} \mathbf{1}_n^{(j)},
\end{equation}
thereby avoiding explicit 3D intersection computation.
Since the candidate set grows over iterations, retrieval cost increases accordingly. Thus, it makes the GPU approximation much more efficient for long iterative generation while preserving geometry-aware ranking.

\subsection{Large Model Scaling Up}
\label{sec:large_model}
Inspired by LingBot-World~\cite{team2026advancing}, we further perform scale-up training on a MoE-28B backbone to improve generalization, dynamic modeling capability, and long-horizon consistency.
In contrast to this, we observe that training the action module only in the high-noise model is sufficient for achieving precise control, while the low-noise model can be trained independently of the action module to focus on refining visual details. Accordingly, we train the high-noise model using action-accurate data, where the relatively narrow noise regime allows the action control to converge efficiently. Meanwhile, the low-noise model can be trained with Internet video data to improve generalization. By decoupling the capabilities of action control and refinement of visual quality to different staged models, the potential of unlabeled data can be largely leveraged. 

During training, we progressively scale both the resolution and video clip length, which accelerates convergence and stabilizes long-horizon behavior.
Furthermore, since first-person and third-person dynamics are difficult to model jointly, we adopt a viewpoint-specialized design. Specifically, we train two separate high-noise models for first-person and third-person views, respectively, while sharing a common low-noise model. This viewpoint-specific specialization enables efficient resource allocation and allows the model to support both immersive first-person experiences and third-person game-oriented scenarios.
For minute-level long-horizon generation, leveraging the scale of MoE-28B, we focus on generating high-fidelity, minute-long video sequences with enhanced long-term memory retention. This significantly improves frame-to-frame temporal consistency and context preservation, bridging the gap between short-clip generation and longer, narrative-style video generation.
\section{Data System}
We build a multi-source data system for world model training, integrating Unreal Engine-based rendering, gameplay recording for dynamic third-person data, real-world video datasets, and unified annotation and filtering. The pipeline prepares video data spanning static and dynamic scenes across multiple viewpoints.
\begin{figure}[!htb]
    \centering
    \includegraphics[width=0.99\textwidth]{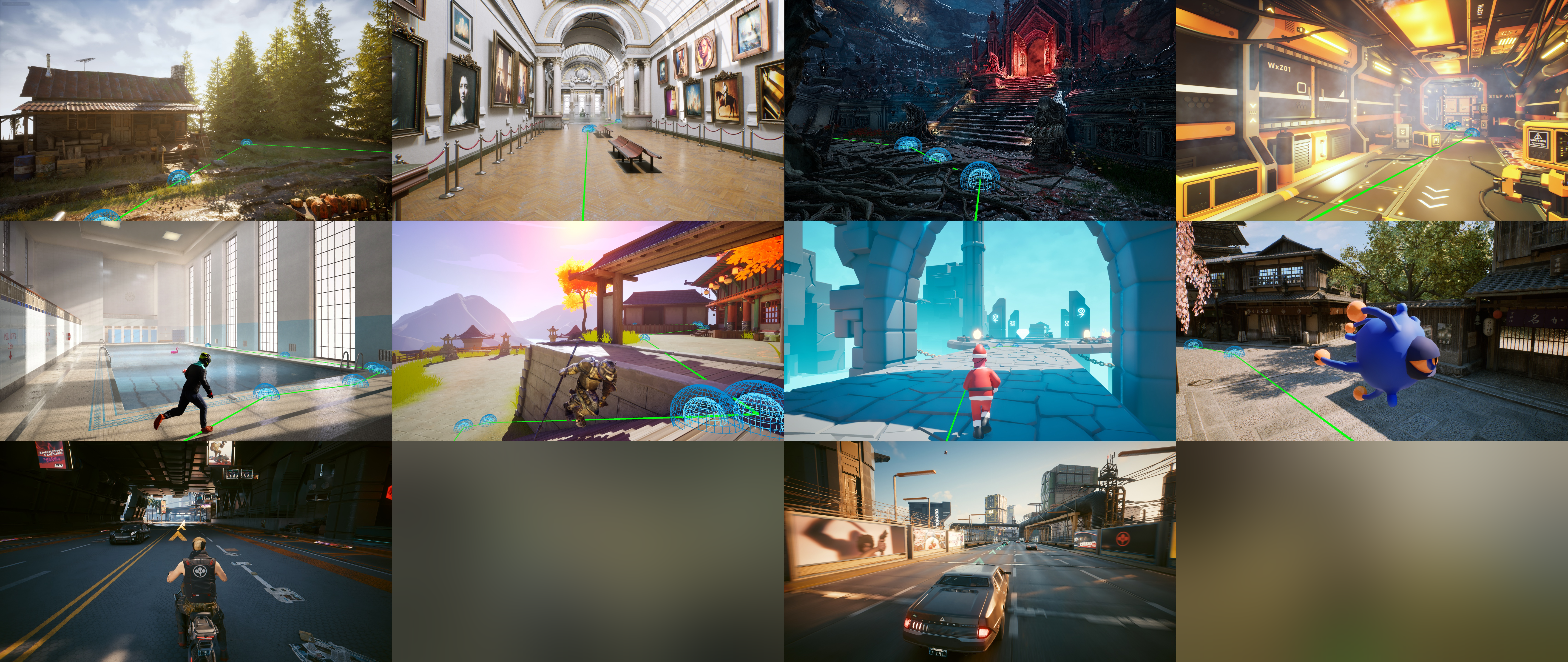}
    \caption{\textbf{Representative rendered and recorded scenes and agent trajectories from our data pipeline.} The panels in row 3, columns 2 and 4 (counting rows from top to bottom and columns from left to right), are blurred for publication-related redaction.}
    \label{fig:all_data_demo}
\end{figure}
\subsection{Unreal Engine-based Data Production}

Our Unreal Engine-based pipeline---Unreal-Gen---produces video from more than 1,000 UE5 scenes built on Nanite virtualized geometry and Lumen global illumination. The core design principle is \emph{tick-level synchronization}: in each rendered frame $t$, the system simultaneously captures:
\begin{equation}
\mathcal{D}_t = \left(\mathbf{I}_t,\ \mathbf{p}_t,\ \mathbf{r}_t,\ \mathbf{c}_t,\ \boldsymbol{\theta}_t,\ \mathbf{a}_t\right)
\end{equation}
where $\mathbf{I}_t \in \mathbb{R}^{H \times W \times 3}$ is the rendered RGB frame, $(\mathbf{p}_t, \mathbf{r}_t)$ the player's world position and rotation, $(\mathbf{c}_t, \boldsymbol{\theta}_t)$ the camera's 6-DoF pose, and $\mathbf{a}_t \in \{0,1\}^6$ the discrete action vector. All quantities are sampled within the same engine tick callback to align the recorded video and state information at the engine-tick level. Double-precision quaternion arithmetic is used for rotation calculations.

\textbf{NavMesh--RL Hybrid Agent.}A two-level hierarchy drives autonomous exploration: a high-level RL policy $\pi_\theta(\mathbf{g}_t \mid \mathbf{s}_t)$ selects navigation goals via an intrinsic reward combining coverage bonus and scene richness, while a low-level NavMesh planner computes collision-free paths. Cascaded fallback strategies (directional coverage, shape route generation, multi-radius retry) and triple stuck detection (position delta, path timeout, bounding box) support navigation across open landscapes, dense indoor environments, and stylized worlds.

\textbf{Camera \& Character Diversity.}
Stochastic yaw randomization (full 360° sweep or 8-direction discrete) and pitch randomization operate concurrently with navigation. The system supports both first-person and third-person perspectives. A modular assembly system decomposes each character into swappable components---tops, pants, shoes, hair, outerwear, hats, and accessories---yielding $|\mathcal{C}| = \prod_i |\mathcal{S}_i| > 10^8$ unique variants sampled at runtime, providing combinations of character appearance across recording sessions.

\textbf{Automated Rendering and Monitoring.}
An automated pipeline orchestrates batch rendering (multiple concurrent UE5 instances per GPU with shader warmup), real-time cloud synchronization, and automated quality assurance (identical frame detection, position anomaly repair, incomplete scene detection). Closed-loop monitoring via webhook enables a single operator to oversee dozens of recording machines.

\subsection{Game-Environment Video Recording}

We design a four-layer decoupled architecture for synchronized recording of gameplay video and associated state information across supported game titles.

\textbf{Four-Layer Architecture.}
(1)~\emph{Game Process Layer}---each supported game runs as an independent process with an injected, game-specific agent plugin for character control and state extraction.
(2)~\emph{Agent Layer}---NavMesh-based autonomous exploration with multi-state-machine coordination and 8-direction viewpoint switching, exporting per-frame state $\mathbf{s}_t = \{\mathbf{p}_t, \mathbf{r}_t, \mathbf{c}_t, \boldsymbol{\theta}_t, a_{\text{nav}}, a_{\text{jump}}, a_{\text{attack}}\}$ via JSON-file IPC ($<$1\,ms polling).
(3)~\emph{Recording Coordination Layer}---OBS Studio segmented recording (60\,s segments) via WebSocket, with physics-based WSAD inference: position deltas are projected onto the camera's local coordinate frame,
\begin{equation}
\mathbf{f} = (\cos\theta_{\text{yaw}},\ \sin\theta_{\text{yaw}}), \quad
\mathbf{r} = (\sin\theta_{\text{yaw}},\ -\cos\theta_{\text{yaw}})
\end{equation}
and classified into eight cardinal directions based on $\langle \Delta\mathbf{p}_t, \mathbf{f} \rangle$ and $\langle \Delta\mathbf{p}_t, \mathbf{r} \rangle$, providing action labels inferred from recorded motion.
(4)~\emph{Dataset Output Layer}---MP4 video paired with per-frame CSV containing six-dimensional action vectors, camera parameters, and full pose information.

\textbf{Scalability \& Reliability.}
The architecture separates game-specific integration in the Agent Layer from the recording and output layers, facilitating reuse across supported titles. Reliability measures include stuck detection with automatic recovery, segmented recording with reconnection, and remote monitoring via webhook alerts. Where supported by the game-specific integration, environment parameters such as weather, time of day, and NPC density are randomized across recording sessions. Overall data accuracy exceeds 99\%.

\subsection{Real-World Data}
\label{sec:real-world-data}

To complement synthetic data in photometric variation and natural camera trajectories, we incorporate four real-world video datasets that collectively span static architectural interiors, large-scale outdoor landmarks, urban street-level locomotion, and diverse aerial and vehicular perspectives.

DL3DV-10K~\cite{ling2023dl3dv10klargescalescenedataset} comprises over 10,000 4K video sequences across 65 point-of-interest categories.
RealEstate10K~\cite{zheng2025realcamvidhighresolutionvideodataset} provides indoor real-estate walkthroughs with purely static scenes and exceptionally clean camera trajectories.
OmniWorld-CityWalk~\cite{zhou2025omniworld} aggregates first-person urban walking footage from YouTube under diverse weather and lighting conditions, with relative poses estimated via DPVO~\cite{teed2023deeppatchvisualodometry}.
SpatialVid-HD~\cite{wang2025spatialvid}, the largest subset, covers pedestrian, driving, and drone-aerial scenarios in high definition, substantially improving coverage of rare scene types and long-tail viewpoint distributions.
Although some of these datasets ship with bundled pose annotations, we uniformly re-annotate all real-world data using ViPE~\cite{huang2025vipe} to standardize coordinate conventions and pose representations across sources.

\subsection{Data Annotation and Quality Filtering}
\label{sec:annotation}
\textbf{Textual Annotation and Quality Assessment.}
We generate fine-grained captions across the synthetic, recorded-game, and real-world video subsets. Specifically, every clip is annotated with structured descriptions produced by InternVL3.5-8B~\cite{wang2025internvl3_5}. Following the description framework of LingBot World~\cite{robbyantteam2026advancingopensourceworldmodels}, we adopt a four-tier hierarchical schema:
(i)~\emph{narrative captions} supply holistic semantic summaries;
(ii)~\emph{static scene captions} decouple scene appearance from camera motion for appearance-conditioned modelling;
(iii)~\emph{dense temporal captions} provide per-segment event labels and camera-motion descriptions; and
(iv)~\emph{perceptual quality scores} rate each clip on a 0--10 scale along five dimensions---motion smoothness, background dynamics, scene complexity, physics plausibility, and overall quality.

\textbf{Trajectory and Speed Filtering.} To address residual pose errors and abnormal motion, we apply three complementary filters: local geometric consistency via depth reprojection error, global motion anomaly via the max-to-median displacement ratio, and camera speed filtering based on median velocity (removing slower and faster clips). A trajectory is retained only if it satisfies all three criteria. Thresholds are calibrated on UNREAL ground truth and applied on per dataset. Combined with perceptual quality filtering, the pipeline removes ~20\% of the raw data, yielding a high-quality curated training set.

\section{Experiments}
\subsection{Setting}
\noindent\textbf{Interactive Base Model.}
Our base model is built upon the Wan2.2-TI2V-5B \cite{wan2025wan} architecture, with action modules integrated into the first 15 DiT blocks, following the design in Matrix-Game 2.0. During training, we jointly optimize cases that include the VAE reference latent (corresponding to the first frame of input videos). With a probability of 0.8, the model is trained using a combination of 4 past-frame latents and 10 current-frame noisy latents. Otherwise, both the past-frame and memory latents are masked out, reducing the task to an action-conditioned image-to-video (I2V) setting. This training mode corresponds to the first segment in practical streaming inference, where the model takes the reference latent together with 14 current-frame noisy latents as input. We then fine-tune the model with a learning rate of $2 \times 10^{-5}$ for 50K steps.

\noindent\textbf{Memory-augmented Base Model.}
To study the effect of the proposed memory design, we initialize the checkpoint from the same action-modulated base model unless otherwise specified.
Inspired by the multi-head RoPE jitter idea in LoL~\cite{cui2026lollongerlongerscaling}, we introduce the head-wise perturbed rotary base during both training and inference stages, allowing the model to learn under the modified temporal encoding throughout optimization. In practice, the perturbation coefficients $\epsilon_h$ are linearly spaced across attention heads, with $\sigma_\theta$ fixed at 0.8, so that each head is associated with a distinct effective rotary base $\hat{\theta}_h$.
During training, we jointly model 5 memory latents, 4 past-frame latents, and 10 noisy latents to be generated, concatenating them before sending into the DiT. The training set contains about 4.8M video clips.

\noindent\textbf{Distillation Model.}
For distillation, the models (teacher, critic, and student) are all directly initialized from the memory-augmented base model. To prevent the few-step student from collapsing during multi-segment inference at first, we employ a cold-start stage under a single-segment inference setting, where ground truth clips are used as the past frames. In this stage, the learning rates of the student and critic model are set to $5 \times 10^{-7}$ and $1 \times 10^{-7}$, respectively. And the student is updated for 5 steps per iteration. The cold-start stage spans the first 600 training steps. Subsequently, we switch to a multi-segment inference setting to adapt the practical streaming use cases. The number of segments $k$ under this setting is randomly sampled from 1 to 6 in each iteration. During this stage, both the student and critic use a learning rate of $1 \times 10^{-7}$, and the student is updated for 3 steps per iteration. This stage is trained for 2,400 steps in total. Moreover, consistent with the base-model setting, the past frames and memory are masked with a probability of 0.2.

\begin{figure}[htbp]
    \centering
    \includegraphics[width=1\textwidth]{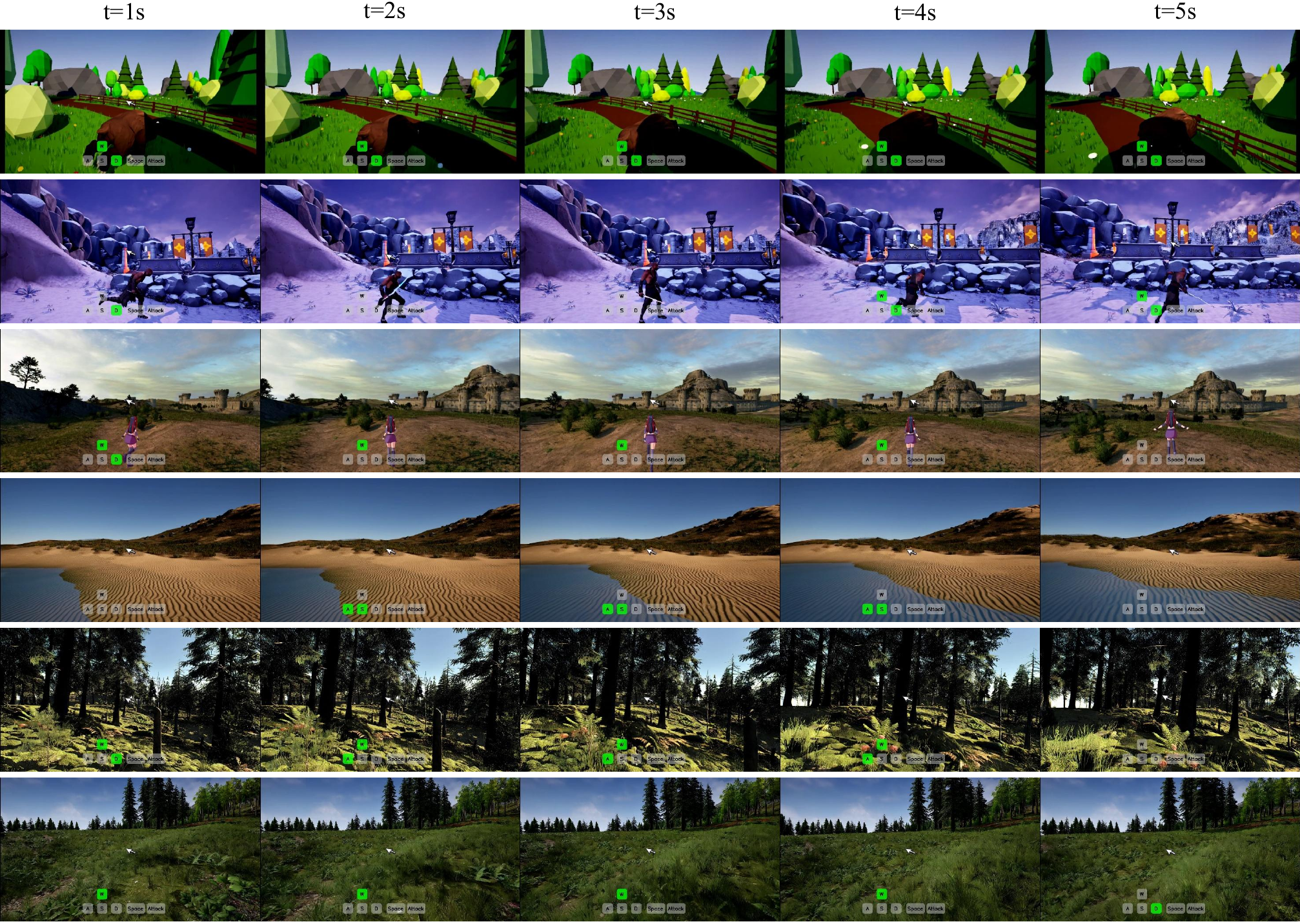}
    \caption{Qualitative results of our interactive base model. The action symbol denotes the action of the current frame.}
    \label{fig:result_basemodel}
\end{figure}

\subsection{Results and Analysis}
\subsubsection{Base Model}
Figure~\ref{fig:result_basemodel} demonstrates the capability of our interactive base model. As shown in the figure, the character exhibits basic controllability, while the background remains stable without noticeable drift. Moreover, as the camera moves, the scene content shows a reasonable zoom-in and zoom-out relationship that is consistent with the camera motion.

\begin{figure}[!h]
    \centering
    \includegraphics[width=1\textwidth]{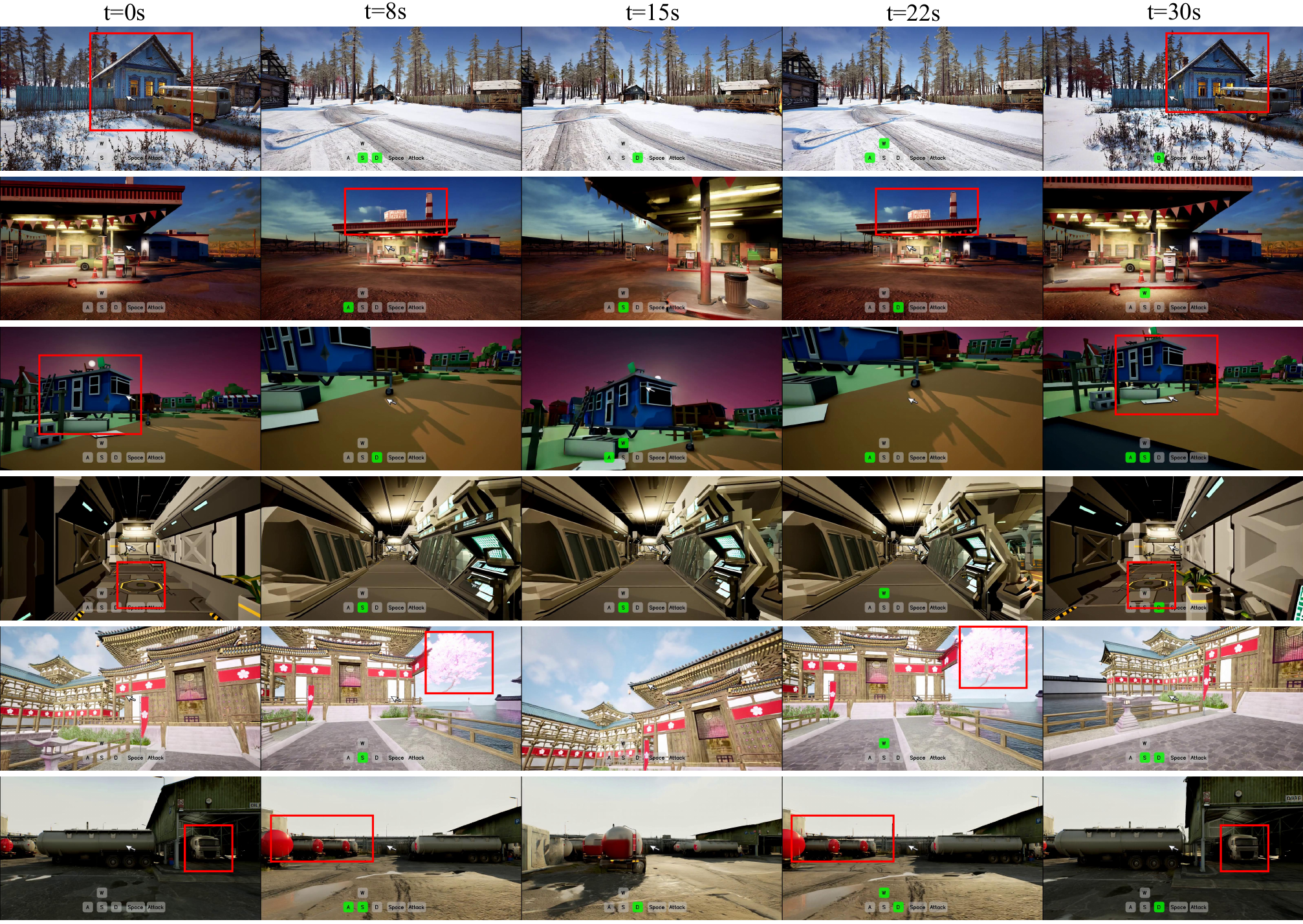}
    \caption{Memory-based scene revisitation in long videos. Each row is sampled uniformly in time; the first frame is the input image, and the second-half actions reverse the first-half actions.}
    \label{fig:memory-result}
\end{figure}

Figure~\ref{fig:memory-result} evaluates memory-augmented base model under a controlled scene-revisitation setting. Specifically, each example is a rollout sampled at uniform temporal intervals, where the first frame is the input image and the second-half actions reverse those in the first half, forcing the camera to return to previously visited regions. In this setting, successful reconstruction during the return phase cannot be explained by short-term continuity alone, but requires effective use of long-range memory. As shown in the figure, when revisiting earlier viewpoints, the model can faithfully recover previously observed scene structures and fine-grained appearance details, including local geometry, object configuration, facade patterns, and texture-level cues highlighted by the red boxes. This behavior is consistent with our memory design: camera-aware retrieval selects view-relevant history, while unified self-attention and relative geometric encoding allow the model to reuse retrieved memory in a structured and view-consistent manner. The result suggests that the proposed memory mechanism provides an effective long-range anchor for scene layout and appearance, which is important for maintaining consistency in extended interactive rollouts.

\begin{figure}[!htb]
    \centering
    \includegraphics[width=1\textwidth]{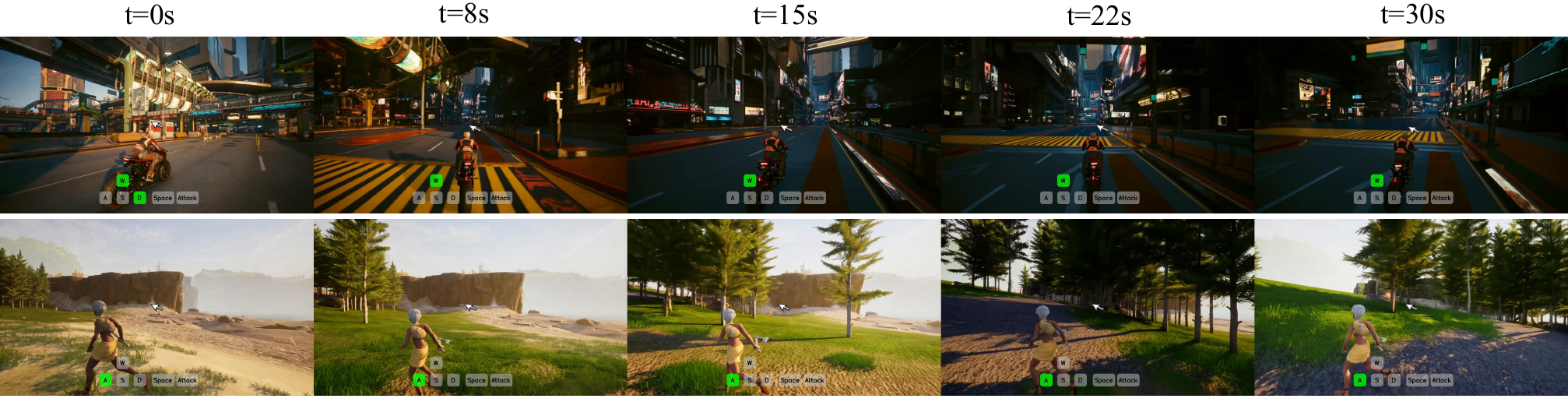}
    \caption{Qualitative results of our 28B model on third-person video generation.}
    \label{fig:third-person-results}
\end{figure}

Figure~\ref{fig:third-person-results} further presents qualitative results of our 28B model on long-horizon third-person video generation, with the displayed examples showing urban motorcycle traversal and outdoor exploration. In these examples, the generated videos exhibit strong temporal consistency in scene layout, character identity, and object relations, while also maintaining vivid motion dynamics under continuous camera and action changes. These examples suggest that scaling the model further improves both the stability and expressiveness of interactive long-video generation, allowing the system to preserve coherent world structure while producing richer motion, lighting variation, and scene transitions across the displayed environments.

\subsubsection{Distillation Model}

\begin{figure}[!htb]
    \centering
    \includegraphics[width=1\textwidth]{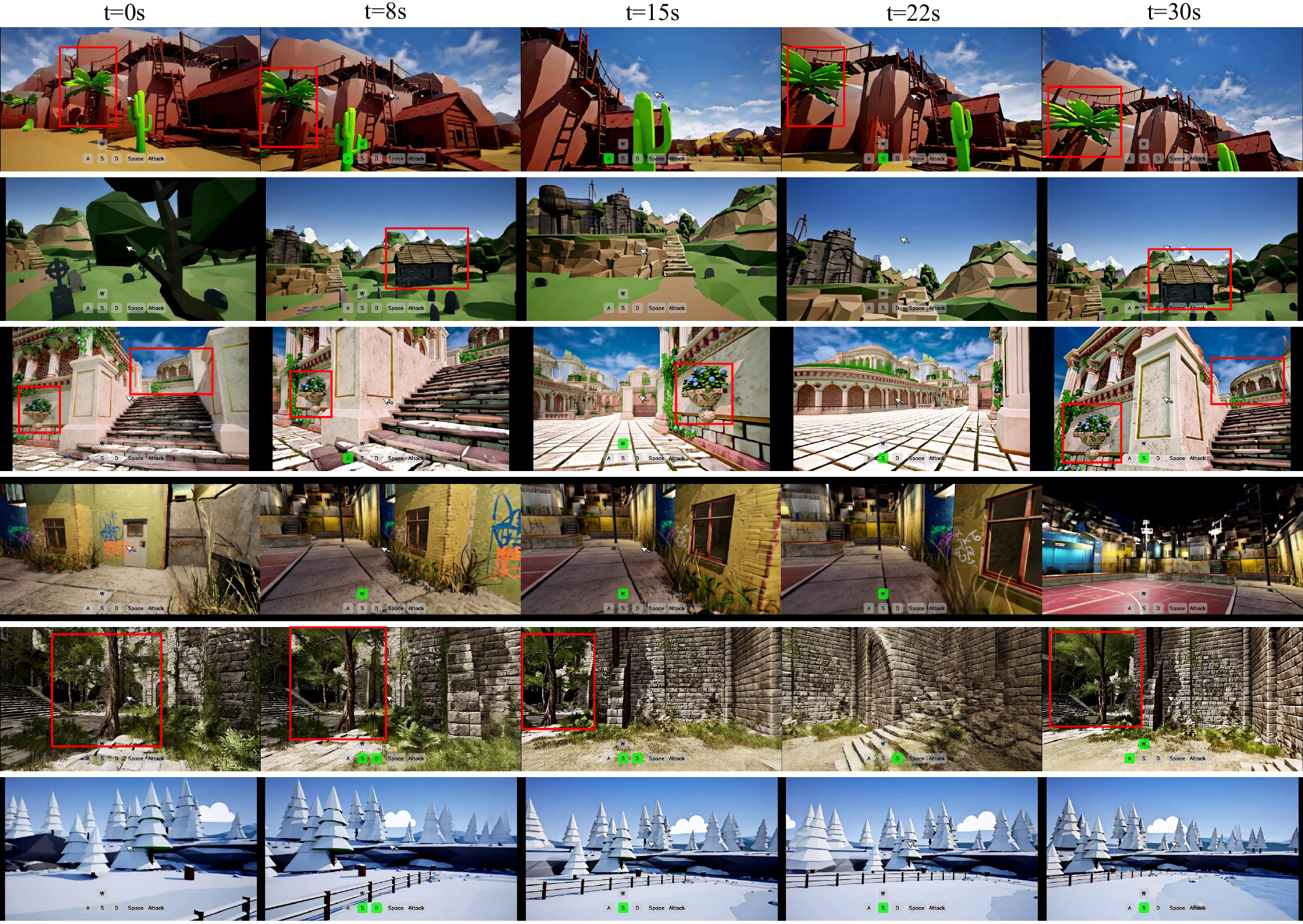}
    \caption{Qualitative results of our distilled model. Each row is sampled uniformly over time.}
    \label{fig:result_dist}
\end{figure}

We further evaluate the distilled model on long-horizon generation. Figure\ref{fig:result_dist} presents the results. We deliberately design the action sequence shown here to revisit several specific viewpoints and scene contents. As can be observed, the distilled model effectively inherits the memory capability of the memory-augmented base model. Content that previously appeared in the scene and was later occluded can be faithfully reproduced in subsequent frames. In addition, the model also demonstrates rich and accurate generation ability for newly emerged scenes, with no noticeable drift in style or content in the later stages of the sequence.

\subsubsection{Real-Time Inference}

\begin{table}[h]
\centering
\small
\setlength{\tabcolsep}{5pt}
\caption{\textbf{Ablation on major acceleration components with 75\% VAE pruning.} We report the FPS change after removing each major acceleration component from the full inference setup.}
\label{tab:ablation_h200_async}
\begin{tabular}{lcc}
\toprule
\textbf{Configuration} & \textbf{FPS} & \textbf{$\downarrow$ Drop} \\
\midrule
Full & $\thicksim$40 & -- \\
- INT8 quantization & 27.38 & 12.62 \\
- MG-LightVAE & 25.79 & 14.21 \\
- GPU retrieval & 6.60 & 33.40 \\
\bottomrule
\end{tabular}
\end{table}

We conduct ablation studies on several key acceleration strategies, including INT8 quantization for DiT, VAE pruning, and GPU-based memory retrieval. Unless otherwise specified, all experiments are performed under an asynchronous 8+1 setup, with 8 GPUs for DiT inference and 1 GPU for VAE decoding. We report throughput on both H-series and A-series GPUs, where FlashAttention 3 and 2 are adopted, respectively. Together, these techniques improve the efficiency of the full pipeline and enable a final inference speed of up to 40 FPS. Under a single-node 7+1 setup, the throughput is slightly lower, while the overall conclusions remain unchanged.

From Table~\ref{tab:ablation_h200_async}, the full configuration demonstrates the highest throughput, showing that each acceleration component effectively contribute to the final speedup. INT8 quantization and MG-LightVAE both provide clear efficiency gains, while GPU retrieval is the most critical component, as removing it causes the largest FPS drop. Overall, the final acceleration comes from the combined effect of multiple optimizations rather than any single technique alone. More broadly, H-series GPUs also consistently deliver higher throughput than A-series GPUs under the same parallel setting.

\begin{figure}[!htb]
    \centering
    \includegraphics[width=1\textwidth]{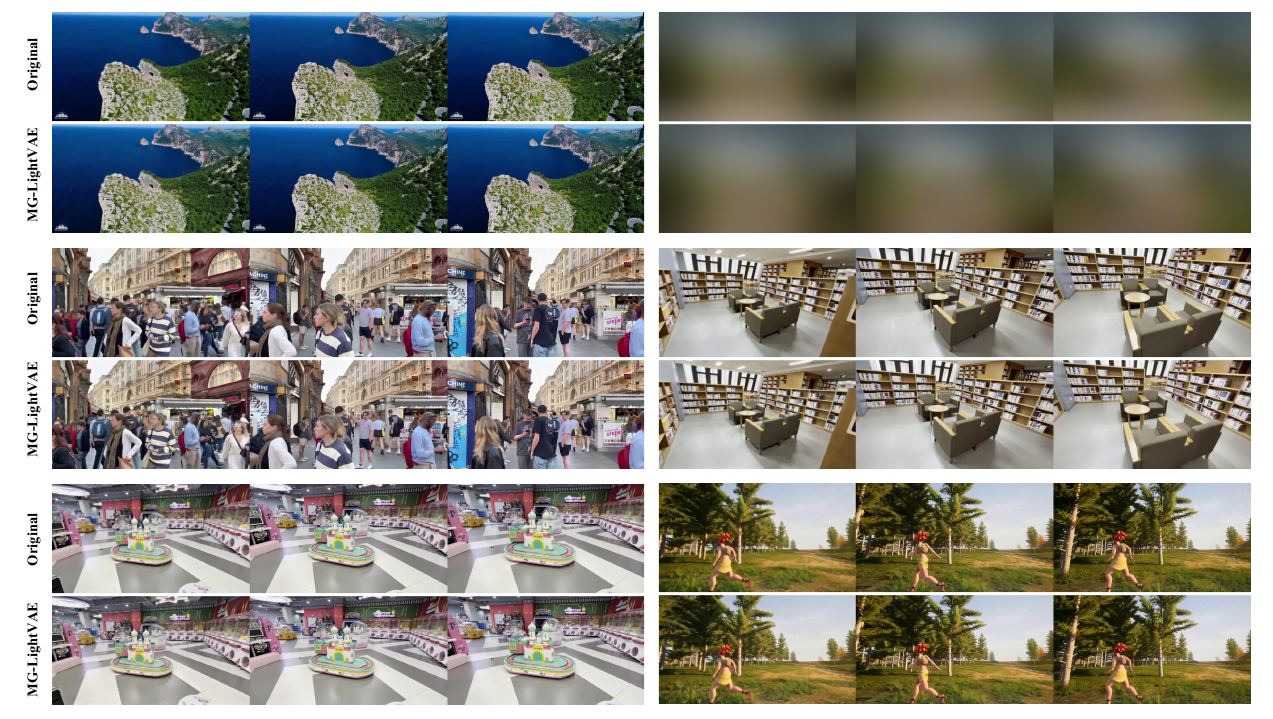}
    \caption{For each case, the top row shows the original video and the bottom row shows the reconstruction by the 50\% pruned MG-LightVAE. The upper-right original/reconstruction example is blurred for publication-related redaction; these panels are not used for visual comparison.}
    \label{fig:vae-recon}
\end{figure}

After optimizing the inference speed of DiT, VAE decoding latency becomes the main bottleneck in streaming high-resolution generation. To address this, we train two pruned variants of MG-LightVAE with 50\% and 75\% pruning ratios. The evaluation is conducted from two aspects: reconstruction quality and inference efficiency. For quantitative results, we use a mini test dataset and compare the reconstructions of the original Wan2.2 VAE, the 50\% pruned MG-LightVAE, and the 75\% pruned MG-LightVAE against the original videos using PSNR and SSIM. We also measure the reconstruction time on 17-frame videos at 720$\times$1280 resolution, and report both the full reconstruction time (encoder+decoder) and the decoder-only time. As shown in Table~\ref{tab:vae_psnr_ssim}, the 50\% pruned MG-LightVAE maintains strong reconstruction quality with only a limited drop compared with the original Wan2.2 VAE, while substantially reducing inference time. The 75\% pruned version provides further speedup at the cost of a larger degradation in reconstruction fidelity, but still preserves high similarity overall. Qualitative examples are shown in Figure~\ref{fig:vae-recon}. The 50\% pruned MG-LightVAE preserves the main scene structure and visual content well.

\begin{table}[t]
\centering
\caption{Reconstruction quality and efficiency comparison between the original Wan2.2 VAE and MG-LightVAE with 50\% and 75\% pruning. \textbf{Full} denotes encoder+decoder time, and \textbf{Dec.} denotes decoder-only time. Higher PSNR and SSIM indicate better reconstruction fidelity.}
\begin{tabular}{lcccc}
\toprule
\textbf{Model} & \textbf{PSNR} $\uparrow$ & \textbf{SSIM} $\uparrow$ & \textbf{Full(s)} $\downarrow$ & \textbf{Dec.(s)} $\downarrow$ \\
\midrule
Wan2.2 VAE                & 33.79 & 0.99 & 0.99 & 0.76 \\
MG-LightVAE (50\% pruned) & 31.84 & 0.99 & 0.52 & 0.30 \\
MG-LightVAE (75\% pruned) & 31.14 & 0.99 & 0.35 & 0.13 \\
\bottomrule
\end{tabular}
\label{tab:vae_psnr_ssim}
\end{table}

\section{Conclusion}
We present Matrix-Game 3.0, a unified framework for interactive world modeling that jointly achieves long-horizon consistency, high-resolution generation, and real-time inference. 
Our approach adopts a co-design across data, modeling, and system deployment: a multi-source data preparation system for interactive supervision; an error-aware interactive base model for robust self-correction under iterative generation; a camera-aware memory mechanism for long-horizon spatiotemporal consistency; and a training–inference aligned multi-segment distillation framework with system-level optimizations for real-time performance.
Experimental results show that Matrix-Game 3.0 achieves up to 40 FPS real-time generation at 720p resolution with a 5B model, while maintaining stable spatiotemporal consistency over minute-long sequences. Scaling to larger models further improves generation quality, dynamic behavior, and generalization capability.
Future directions include scaling model and data for improved generation quality, developing more efficient architectures for higher resolution and longer sequences, and exploring more advanced memory mechanisms for better long-term dependency modeling and complex interaction.

{
\small
\bibliography{neurips_2025}
\bibliographystyle{plain}
}

\end{document}